%% file: main.tex
\documentclass[twoside,11pt]{article}

\usepackage{jmlr2e}
\usepackage{xcolor}
\usepackage[frozencache,cachedir=./]{minted}
\usepackage{tcolorbox}
\usepackage{amsmath}
\usepackage{mathtools}
\input{tikzsetup}

\usepackage{lastpage}
\jmlrheading{23}{2024}{1-\pageref{LastPage}}{3/22; Revised
7/22}{11/22}{22-0277}{D\"urholt, Asche, Kleinekorte, Mancino-Ball, Schiller, Keupp, Sung, Osburg, Boyne, Misener, Eldred, Steuer Costa, Kappatou, Lee, Linzner, Walz, Wulkow, Shafei}

\newcommand{\bofire}{\texttt{BoFire}}

\mathtoolsset{showonlyrefs}

\NewTColorBox{realworld}{ O{red} m d"" !O{} }
{colback=white!95!black,colframe=white!75!black,arc=0pt,outer arc=0pt,top=0pt,bottom=0pt,left=5mm,right=5mm,
leftrule=0pt,rightrule=0pt,toprule=0.3mm,bottomrule=0.3mm,boxsep=0.5mm,}

\NewTotalTCBox{\realworldinline}{ O{red} v !O{} }
{nobeforeafter,tcbox raise base,arc=0pt,outer arc=0pt,
top=0pt,bottom=0pt,left=0mm,right=0mm,
leftrule=0pt,rightrule=0pt,toprule=0.3mm,bottomrule=0.3mm,boxsep=0.5mm,
colback=white!95!black,colframe=white!75!black,#3}{#2}

\ShortHeadings{Bayesian Optimization Framework Intended for Real Experiments}{D\"urholt, et al.}
\firstpageno{1}

\begin{document}

\title{BoFire: Bayesian Optimization Framework Intended for Real Experiments}

% see https://www.jmlr.org/papers/volume24/22-0809/22-0809.pdf
% for a more condensed author list

\author{\name Johannes P. D\"urholt$^{1}$, Thomas S. Asche$^{1}$, Johanna Kleinekorte$^{1}$, Gabriel Mancino-Ball$^{1}$, Benjamin Schiller$^{1}$, Simon Sung$^{1}$, Julian Keupp$^{2}$, Aaron Osburg$^3$, Toby Boyne$^{4}$, Ruth Misener$^{4}$, Rosona Eldred$^{5}$, Wagner Steuer Costa$^{5}$, Chrysoula Kappatou$^{6}$, Robert M. Lee$^{6}$, Dominik Linzner$^{6}$, David Walz$^{6}$, Niklas Wulkow$^{6}$, Behrang Shafei$^{6}$ 
\\ \email{johannespeter.duerholt@evonik.com}
\\
       \addr 
       $^1$Evonik Operations GmbH, DE, 
       $^2$Boehringer Ingelheim Pharma GmbH \& Co. KG, DE,
       $^3$Heidelberg University, DE,
       $^4$Imperial College London, UK,
       $^5$Chemovator GmbH, DE,
       $^6$BASF SE, DE
       }

\editor{}

\maketitle

\begin{abstract}
Our open-source Python package BoFire combines Bayesian Optimization (BO) with other design of experiments (DoE) strategies focusing on developing and optimizing new chemistry. Previous BO implementations, for example as they exist in the literature or software, require substantial adaptation for effective real-world deployment in chemical industry. BoFire provides a rich feature-set with extensive configurability and realizes our vision of fast-tracking research contributions into industrial use via maintainable open-source software. Owing to quality-of-life features like JSON-serializability of problem formulations, BoFire enables seamless integration of BO into RESTful APIs, a common architecture component for both self-driving laboratories and human-in-the-loop setups. This paper discusses the differences between BoFire and other BO implementations and outlines ways that BO research needs to be adapted for real-world use in a chemistry setting.
\end{abstract}

\begin{keywords}
  Bayesian optimization, Design of experiments, Active learning
\end{keywords}

\section{Introduction}

Once a chemist has outlined a possible reaction for creating a new chemical, or proposed a formulation or process for a new product, the focus in industrial chemistry shifts towards optimization. There are lots of questions that need to be answered. For example: \emph{How, by changing the temperature and pressure of the reaction, can we maximize the yield and the purity of the desired chemical? How, by changing the chemical formulation, can we minimize environmental impact and maximize safety?
%How, by changing the reaction time, can we maximize safety in a production setting? 
Given a set of thousands of candidate molecules, which should be tested in the laboratory when only limited resources are available?}

To answer these questions, the most common approach in industry is still human intuition, trial-and-error, or expensive mechanistic models. However, Bayesian optimization (BO) and design of experiments (DoE) offer great possibilities to the chemical industry: treating chemical experiments as black-box functions and optimizing them in the most efficient manner or uncovering the sources of variation under relevant conditions, respectively \citep{coley2017prediction, hase2018phoenics, shields2021bayesian, THEBELT2022117469, frazier2018bayesian}.

Software tools have been introduced to enhance the application of BO, for instance Ax \citep{bakshy2018ax} and BayBE \citep{fitzner2022baybe}, building on foundational machine learning software like
%PyTorch \citep{paszke2019pytorch} and 
BoTorch \citep{balandat2020botorch}.
%, and Dragonfly \citep{kandasamy2020dragonfly}
The BO tools are complemented by software with cheminformatics capabilities, for example providing representations of molecules, such as SMILES
\citep{RDKit,moriwaki2018mordred,griffiths2023neurips}.

However, in an industrial chemistry setting, existing BO and active learning software would require substantial adaptation prior to deployment. 
Further, as experiments grow in scale and complexity, coordinating between lab components becomes challenging: 
inconsistent data handling makes implementing standalone software into a larger pipeline infeasible.
Following the needs in chemical industry, we have developed (and continue to develop) the open-source software package \emph{Bayesian Optimization Framework Intended for Real Experiments} or \bofire\footnote{\url{https://github.com/experimental-design/bofire}}. 
Our companies actively deploy \bofire{} in both self-driving labs and human-in-the-loop applications.  \bofire{} also supports serialization, whereby all of its components can be translated into a RESTful format, providing an API out of the box and simplifying implementation in existing systems. By making the algorithmic component of our industrial software open-source, we seek to give machine learning researchers a path towards fast-tracking their research ideas into practice and to provide an easy to use tool for practitioners in chemical industry.

\paragraph{Comparison to related work.} 
The frameworks most similar to \bofire{} are Ax \citep{Chang2019arXiv} and
BayBE \citep{fitzner2022baybe}. Compared to Ax, \bofire{} offers chemoinformatics capabilities, classical DoE approaches and serialization via Pydantic \citep{colvin2024pydantic}. Compared to BayBE, \bofire{} offers DoE strategies, serialization via Pydantic, and other application-relevant features such as true multi-objective optimization compared to a pure scalarization based approach.
We developed \bofire{} to meet the BO and DoE needs of industrial chemists in a single package.

\section{Integrating experimental design into real-world labs}
\label{scn:realworldlabs}

\begin{figure}[!t]
    \centering
    \input{fig_bofire_flowchart}
    \caption{\bofire{} provides a complete interface for defining and solving optimization problems in the lab. All objects in the loop - candidates, strategies, surrogates, and proposals - are fully serializable.}
    \label{fig:bofire_flow}
\end{figure}
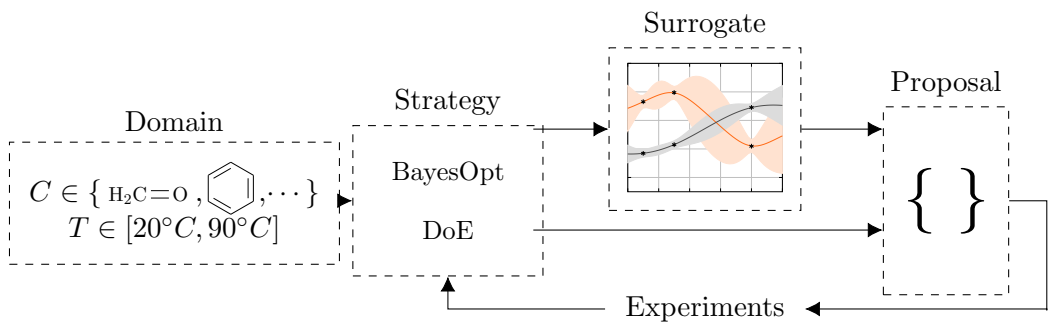

We take an experimentalist-first approach to the software architecture, implementing features that are industrially useful and focusing on easy user deployment. A \realworldinline{real-world example} motivates this section (with corresponding code in Listing \ref{lst:data_model} and visualization in Figure \ref{fig:bofire_flow}) and our GitHub repository features other examples in Jupyter notebooks.

\paragraph{Domains.}
In \bofire{}, a \texttt{Domain} consists of \texttt{Inputs}, \texttt{Outputs}, and optionally \texttt{Constraints}.
\bofire{} allows the user to define an input space $\mathcal{X} = x_1 \otimes x_2 \ldots \otimes x_D$ where the input features $x_i$ can be continuous, discrete, molecular or categorical.
%Input features also support data transformations, such as one-hot encoding and SMILES strings for molecules.
\bofire{} supports the following constraints: (non)linear (in)equality, NChooseK, and interpoint equality. The package also provides support for learning black-box inequality constraints.
\begin{realworld} .
    A chemist designs a paint using a selection of 20 different compounds, each of which has a continuously-varying concentration. They use an \texttt{NChooseK} constraint to limit each test-paint mixture to at most 5 compounds. For a batch of multiple mixtures, all paints are tested at the same temperature, requiring an \texttt{InterpointEquality} constraint which keeps the temperature fixed during the batch of experiments.
\end{realworld}

\paragraph{Objectives.}
In \bofire{}, objectives are defined separately from the outputs on which they operate.
This allows us to define outputs in a physically meaningful way.
Here, minimization, maximization, close-to-target and sigmoid-type objectives are supported.
For multi-objective optimization, \bofire{} supports two schemes: an \textit{a priori} approach, in which the user specifies an additive or multiplicative weighting of each objective; and an \textit{a posteriori} approach, where the optimizer approximates the Pareto front of all optimal compromises for subsequent decision-making.
The latter is implemented via \texttt{qParEGO} \citep{knowles2006parego} and \texttt{q(log)(N)EHVI} strategies \citep{Daulton2020qEHVI, Daulton2021qNEHVI, Ament2023neurips}. Both can be used in combination with black box constraints.

\begin{realworld} .
    The chemist wants to achieve a target viscosity, while maximising hydrophobicity. They define the measurements as \texttt{Output}s, and use the \texttt{CloseToTargetObjective} and \texttt{MaximizeObjective} respectively to drive the optimization.
\end{realworld}

\paragraph{Strategies.}
Given a \texttt{Domain}, the user selects a \texttt{Strategy} to generate experimental proposals.
Classical DoE based strategies can generate (fractional)-factorial, space-filling (via sobol-, uniform- or latin-hypercube sampling), and D-,E-,A-,G-, or K-optimal designs. Compared to commercial software (e.g. Modde, JMP), \bofire{} supports designs over constrained mixed-type input spaces.
Alternatively, predictive strategies use \texttt{Surrogates} to model the data-generating process and perform BO. Many of these strategies are built on BoTorch \citep{balandat2020botorch} and provide numerous acquisition functions. They are easily extendable  and allow users to define custom strategies and surrogates, for instance as we did with 
ENTMOOT \citep{thebelt2021entmoot}.

\begin{realworld} .
    The initial paint experiments should be selected using a \texttt{SpaceFillingDesign}, then use a \texttt{PredictiveStrategy} to suggest optimal experiments. The chemist uses the \texttt{StepwiseStrategy} interface to seamlessly transition between strategies.
\end{realworld}

\section{Library Philosophy}

\paragraph{Fully serializable.}
\bofire{} is industry-ready for self-driving labs. In this setting, communication is key: many systems pass data and information between each other, and data integrity is essential.

\bofire{} is natively usable with a RESTful Application Programming Interface (API) and structured json-based, document-oriented databases, via the use of the popular data-validation library Pydantic allowing for seamless integration into FastAPI \citep{ramirez2024fastapi}.

We separate all \texttt{Strategies} and \texttt{Surrogates} into data models, and functional components. Data models are fully json-(de)serializable classes built on Pydantic, which hold complete information regarding the search space, surrogates and strategies. 
This clear distinction allows for a minimal \bofire{} installation consisting only of the data models. This is especially useful in scenarios where a process orchestration layer (POL) is involved as the middle layer between a centrally deployed planner using \bofire{}, and closed-loop equipment. 

\begin{listing}
\begin{minted}{python}
compounds = [f"compound_{i}" for i in range(20)]
domain = Domain.from_lists(
    [ContinuousInput(key="temp", bounds=[20, 90], unit="°C"),
        *(ContinuousInput(key=comp, bounds=[0, 1]) for comp in compounds)],
    [ContinuousOutput(key="viscosity", 
        objective=CloseToTargetObjective(target_value=0.5, exponent=2)),
    ContinuousOutput(key="hydrophobicity", objective=MaximizeObjective())],
    [NChooseKConstraint(
            features=compounds, min_count=1, max_count=5, none_also_valid=False),
        InterpointEqualityConstraint(feature="temp")])
\end{minted}
\caption{Defining the domain of the paint problem in Section \ref{scn:realworldlabs}.}
\label{lst:data_model}
\end{listing}

\paragraph{Modularization.}
\bofire{} is both easy to use and highly customizable with respect to its strategies and surrogates. Each component of \bofire{} is modular - problem definitions are independent of the strategies used to solve them, which are in turn independent of the surrogates used to model the observed data. This separation of responsibility enables a `plug-and-play' approach. By building \bofire{} using the BoTorch library, we can leverage the wide range of software written in the BoTorch ecosystem.

\section{Discussion \& Conclusion}

This paper has presented \bofire, our open-source BO and DoE python package. Representing several companies in the chemical industry, we deploy \bofire{} daily to bring BO and DoE into our companies. Each individual contributing company could have easily developed their own bespoke package, but we joined forces to create \bofire{} because of our vision of catalyzing machine learning research. \bofire{} exemplifies our collaboration goals with researchers, for example those working in academia, for example current work on practical multi-fidelity modeling \citep{bonilla2007multi,folch2023combining}. Through \bofire{}, we offer the possibility for researchers to use our platform to translate new strategies and surrogates into practice. 

\vskip 0.2in
\bibliography{bofire,bofire_software}

\end{document}

%% file: tikzsetup.tex
\usepackage{chemfig}
\usepackage{gensymb}

\usepackage{listings}
\DeclareFixedFont{\ttb}{T1}{txtt}{bx}{n}{9} % for bold
\DeclareFixedFont{\ttm}{T1}{txtt}{m}{n}{9}  % for normal
\DeclareFixedFont{\ttbs}{T1}{txtt}{bx}{n}{9} % for bold
\DeclareFixedFont{\ttms}{T1}{txtt}{m}{n}{9}  % for normal
\usepackage{color}
\definecolor{deepblue}{rgb}{0,0,0.5}
\definecolor{deepred}{rgb}{0.6,0,0}
\definecolor{deepgreen}{rgb}{0,0.5,0}

\newcommand\pythonstyle{\lstset{
	language=Python,
	basicstyle=\ttm,
	otherkeywords={self},             % Add keywords here
	keywordstyle=\ttb\color{black},
	emph={__init__,as},          % Custom highlighting
	emphstyle=\ttb\color{black},    % Custom highlighting style
	stringstyle=\color{black},
	frame=none,                         % Any extra options here
	showstringspaces=false            % 
}}
\newcommand\pythonstylesmall{\lstset{
	language=Python,
	basicstyle=\ttms,
	otherkeywords={self},             % Add keywords here
	keywordstyle=\ttb\color{black},
	emph={__init__,as},          % Custom highlighting
	emphstyle=\ttbs\color{black},    % Custom highlighting style
	stringstyle=\color{black},
	frame=none,                         % Any extra options here
	showstringspaces=false            % 
}}
\lstnewenvironment{python}[1][]{
	\pythonstyle
	\lstset{#1}
}{}
\newcommand\pythoninline[1]{{\pythonstyle\lstinline!#1!}}
\newcommand\pythonsmall[1]{{\pythonstylesmall\lstinline!#1!}}

\usepackage{tikz,calc}
\usetikzlibrary{shapes}
\usetikzlibrary{shapes.misc, patterns}
\usetikzlibrary{positioning}
\usetikzlibrary{arrows.meta, decorations.markings}
\usetikzlibrary{fit}
\usepackage{pgfplots}
\pgfplotsset{compat=newest}
\usepackage{relsize}
\tikzset{%
    >={Latex[width=2mm,length=2mm]},
    base/.style = {fill=none, font=\sffamily, font=\small},
    dtxt/.style = {base, rectangle, draw=none, fill=none, text centered,
                    minimum width=1cm, minimum height=0.5cm},
    examplestrat/.style = {base, %rectangle split, rectangle split parts=2, 
    draw=none, fill=none, text centered,
                    minimum width=1cm, minimum height=0.5cm, text width=2cm},
    modulebox/.style = {base, rectangle,%
        dashed,
        label={[text width=2cm,align=center]above:#1}}
}
\makeatletter
\tikzset{
    nomorepostaction/.code=\makeatletter\let\tikz@postactions\pgfutil@empty,
    standard/.style={postaction={
    		decoration={
                markings,
                mark=at position 1 with {
                    \arrow[line width=1pt]{stealth}
                } },
            decorate,
            nomorepostaction
        	},
        thick,-, % switch off other arrow tips
        every path/.append style=standard
    }
}
\makeatother

%% file: fig_bofire_flowchart.tex
\begin{tikzpicture}[node distance=8mm]
    %every node/.style={font=\sffamily, font=\large}]
	\def\yShift{13mm}
	\def\yShiftTwo{13mm}
	\def\yShiftThree{12mm}
	\def\xShiftOne{20mm}
	\def\xShiftTwo{50mm}

    % INPUTS
    \node (chemical)[align=center]{
    $C \in \{$%
    \scalebox{0.7}{
    \chemfig[double bond sep=4pt,atom sep=2em]{H_2C(=[8]O)}
    }%
    ,%
    \scalebox{0.5}{
    \chemfig[double bond sep=4pt,atom sep=1.8em]{*6(=-=-=-)}
    }%
    ,%
    $\cdots\}$
    \\%
    $T\in [20\degree C, 90 \degree C]$
    };

    \node (boxproblem)[modulebox={Domain}, fit=(chemical),%
        draw,%
    ]{};

    % STRATEGIES
    % \node (strategy)[right=of boxproblem.east]{Strategy};

    \node (predictive)[examplestrat,above right=0.1cm and .3cm of boxproblem.east,anchor=south west]{%
    BayesOpt
    %\nodepart{second}D-Optimal designs, space filling%
    };

    \node (doe)[examplestrat,below right=0.1cm and .3cm of boxproblem.east,anchor=north west]{%
    DoE
    %\nodepart{second}BoTorch, GPs, Acq, ENTMOOT
    };

    \node (boxstrategy)[%
        modulebox={Strategy},fit=(doe)(predictive),%
        draw,%
        minimum height=2cm%
    ]{};

    % SURROGATES
    \node (surrogates)[scale=.3,right=1cm of boxstrategy.east,anchor=south west]{%
        \input{fig_surrogate}
    };

    \node (boxsurrogates)[%
       modulebox={Surrogate},fit=(surrogates),%
        draw,%
    ]{};

    % EXPERIMENTS
    \node (experiments)[below=1cm of boxsurrogates]{Experiments};
    \node (boxexperiments)[%
        modulebox={},fit=(experiments)
    ]{};

    % PROPOSALS
    \node (json)[right=2.5 cm of boxsurrogates|-boxstrategy]
    {\Huge \{ \}};

    \node (boxproposals)[modulebox={Proposal},fit=(json),
    draw,%
    minimum height=2.5cm]{};
  
  % Specification of lines between nodes specified above
  % with aditional nodes for description 

  \draw[->] (boxproblem.east |- boxstrategy) -- (boxstrategy.west);
  \draw[->] (predictive.east |- boxsurrogates) -- (boxsurrogates.west);
  \draw[->] (boxsurrogates.east) -- (boxproposals.west |- boxsurrogates);
  \draw[->] (doe.east) -- (boxproposals.west |- doe);

  % close the loop
  \draw[->] (boxproposals.east) -|%
  ([shift={(0.5cm,-0.2cm)}]boxproposals.south east)  |-%
  % ([shift={(-0.5cm,0.5cm)}]boxproblem.south west) |-%
  (boxexperiments.east);
  \draw[->] (boxexperiments.west) -|
  (boxstrategy.south);

\end{tikzpicture}

%% file: main.bbl
\begin{thebibliography}{21}
\providecommand{\natexlab}[1]{#1}
\providecommand{\url}[1]{\texttt{#1}}
\expandafter\ifx\csname urlstyle\endcsname\relax
  \providecommand{\doi}[1]{doi: #1}\else
  \providecommand{\doi}{doi: \begingroup \urlstyle{rm}\Url}\fi

\bibitem[Ament et~al.(2023)Ament, Daulton, Eriksson, Balandat, and Bakshy]{Ament2023neurips}
Sebastian Ament, Samuel Daulton, David Eriksson, Maximilian Balandat, and Eytan Bakshy.
\newblock Unexpected improvements to expected improvement for {B}ayesian optimization.
\newblock In \emph{NeurIPS}, volume~36, pages 20577--20612, 2023.

\bibitem[Bakshy et~al.(2018)Bakshy, Dworkin, Karrer, Kashin, Letham, Murthy, and Singh]{bakshy2018ax}
Eytan Bakshy, Lili Dworkin, Brian Karrer, Konstantin Kashin, Ben Letham, Ashwin Murthy, and Shaun Singh.
\newblock Ae: A domain-agnostic platform for adaptive experimentation.
\newblock In \emph{NeurIPS Systems for ML Workshop}, 2018.

\bibitem[Balandat et~al.(2020)Balandat, Karrer, Jiang, Daulton, Letham, Wilson, and Bakshy]{balandat2020botorch}
Maximilian Balandat, Brian Karrer, Daniel~R. Jiang, Samuel Daulton, Benjamin Letham, Andrew~Gordon Wilson, and Eytan Bakshy.
\newblock {BoTorch: A Framework for Efficient Monte-Carlo Bayesian Optimization}.
\newblock In \emph{NeurIPS}, 2020.

\bibitem[Bonilla et~al.(2007)Bonilla, Chai, and Williams]{bonilla2007multi}
Edwin~V Bonilla, Kian Chai, and Christopher Williams.
\newblock Multi-task {G}aussian process prediction.
\newblock \emph{NIPS}, 20, 2007.

\bibitem[Chang(2019)]{Chang2019arXiv}
Daniel~T. Chang.
\newblock Bayesian hyperparameter optimization with {BoTorch}, {GPyTorch} and {Ax}.
\newblock \emph{arXiv:1912.05686}, 2019.

\bibitem[Coley et~al.(2017)Coley, Barzilay, Jaakkola, Green, and Jensen]{coley2017prediction}
Connor~W Coley, Regina Barzilay, Tommi~S Jaakkola, William~H Green, and Klavs~F Jensen.
\newblock Prediction of organic reaction outcomes using machine learning.
\newblock \emph{ACS Central Science}, 3\penalty0 (5):\penalty0 434--443, 2017.

\bibitem[Colvin(2024)]{colvin2024pydantic}
Samuel Colvin.
\newblock Pydantic, June 2024.
\newblock URL \url{https://github.com/pydantic/pydantic}.

\bibitem[Daulton et~al.(2020)Daulton, Balandat, and Bakshy]{Daulton2020qEHVI}
Samuel Daulton, Maximilian Balandat, and Eytan Bakshy.
\newblock Differentiable expected hypervolume improvement for parallel multi-objective {B}ayesian optimization.
\newblock In \emph{NeurIPS}, volume~33, pages 9851--9864, 2020.

\bibitem[Daulton et~al.(2021)Daulton, Balandat, and Bakshy]{Daulton2021qNEHVI}
Samuel Daulton, Maximilian Balandat, and Eytan Bakshy.
\newblock Parallel {B}ayesian optimization of multiple noisy objectives with expected hypervolume improvement.
\newblock In \emph{NeurIPS}, volume~34, pages 2187--2200, 2021.

\bibitem[Fitzner et~al.(2022)Fitzner, {\v S}o{\v s}i{'c}, Hopp, and Lee]{fitzner2022baybe}
Martin Fitzner, Adrian {\v S}o{\v s}i{'c}, Alexander Hopp, and Alex Lee.
\newblock {BayBE} – a {Bayesian} back end for design of experiments, 2022.
\newblock URL \url{https://github.com/emdgroup/baybe}.
\newblock Accessed: 2024-02-22.

\bibitem[Folch et~al.(2023)Folch, Lee, Shafei, Walz, Tsay, van~der Wilk, and Misener]{folch2023combining}
Jose~Pablo Folch, Robert~M Lee, Behrang Shafei, David Walz, Calvin Tsay, Mark van~der Wilk, and Ruth Misener.
\newblock Combining multi-fidelity modelling and asynchronous batch {B}ayesian optimization.
\newblock \emph{Computers \& Chemical Engineering}, 172:\penalty0 108194, 2023.

\bibitem[Frazier(2018)]{frazier2018bayesian}
Peter~I. Frazier.
\newblock A tutorial on {B}ayesian optimization.
\newblock \emph{arXiv preprint arXiv:1807.02811}, 2018.

\bibitem[Griffiths et~al.(2023)Griffiths, Klarner, Moss, Ravuri, Truong, Du, Stanton, Tom, Rankovi{\'c}, Jamasb, Deshwal, Schwartz, Tripp, Kell, Frieder, Bourached, Chan, Moss, Guo, D{\"u}rholt, Chaurasia, Park, Strieth-Kalthoff, Lee, Cheng, Aspuru-Guzik, Schwaller, and Tang]{griffiths2023neurips}
Ryan-Rhys Griffiths, Leo Klarner, Henry Moss, Aditya Ravuri, Sang~T. Truong, Yuanqi Du, Samuel~Don Stanton, Gary Tom, Bojana Rankovi{\'c}, Arian~Rokkum Jamasb, Aryan Deshwal, Julius Schwartz, Austin Tripp, Gregory Kell, Simon Frieder, Anthony Bourached, Alex~James Chan, Jacob Moss, Chengzhi Guo, Johannes~P. D{\"u}rholt, Saudamini Chaurasia, Ji~Won Park, Felix Strieth-Kalthoff, Alpha Lee, Bingqing Cheng, Alan Aspuru-Guzik, Philippe Schwaller, and Jian Tang.
\newblock {GAUCHE}: A library for {G}aussian processes in chemistry.
\newblock In \emph{NeurIPS}, 2023.

\bibitem[Hase et~al.(2018)Hase, Roch, Kreisbeck, and Aspuru-Guzik]{hase2018phoenics}
Florian Hase, Lo{\"\i}c~M Roch, Christoph Kreisbeck, and Al{\'a}n Aspuru-Guzik.
\newblock Phoenics: a {B}ayesian optimizer for chemistry.
\newblock \emph{ACS Central Science}, 4\penalty0 (9):\penalty0 1134--1145, 2018.

\bibitem[Knowles(2006)]{knowles2006parego}
Joshua Knowles.
\newblock {ParEGO}: A hybrid algorithm with on-line landscape approximation for expensive multiobjective optimization problems.
\newblock \emph{IEEE transactions on evolutionary computation}, 10\penalty0 (1):\penalty0 50--66, 2006.

\bibitem[Landrum(2006)]{RDKit}
Greg Landrum.
\newblock {RDKit}: Open-source cheminformatics, 2006.
\newblock URL \url{https://www.rdkit.org}.

\bibitem[Moriwaki et~al.(2018)Moriwaki, Tian, Kawashita, and Takagi]{moriwaki2018mordred}
Hirotomo Moriwaki, Yu-Shi Tian, Norihito Kawashita, and Tatsuya Takagi.
\newblock Mordred: a molecular descriptor calculator.
\newblock \emph{Journal of cheminformatics}, 10\penalty0 (1):\penalty0 1--14, 2018.

\bibitem[Ramírez(2024)]{ramirez2024fastapi}
Ramírez.
\newblock Fast{API}, June 2024.
\newblock URL \url{'https://github.com/tiangolo/fastapi'}.

\bibitem[Shields et~al.(2021)Shields, Stevens, Li, Parasram, Damani, Alvarado, Janey, Adams, and Doyle]{shields2021bayesian}
Benjamin~J Shields, Jason Stevens, Jun Li, Marvin Parasram, Farhan Damani, Jesus I~Martinez Alvarado, Jacob~M Janey, Ryan~P Adams, and Abigail~G Doyle.
\newblock Bayesian reaction optimization as a tool for chemical synthesis.
\newblock \emph{Nature}, 590\penalty0 (7844):\penalty0 89--96, 2021.

\bibitem[Thebelt et~al.(2021)Thebelt, Kronqvist, Mistry, Lee, Sudermann-Merx, and Misener]{thebelt2021entmoot}
Alexander Thebelt, Jan Kronqvist, Miten Mistry, Robert~M Lee, Nathan Sudermann-Merx, and Ruth Misener.
\newblock {ENTMOOT}: A framework for optimization over ensemble tree models.
\newblock \emph{Computers \& Chemical Engineering}, 151:\penalty0 107343, 2021.

\bibitem[Thebelt et~al.(2022)Thebelt, Wiebe, Kronqvist, Tsay, and Misener]{THEBELT2022117469}
Alexander Thebelt, Johannes Wiebe, Jan Kronqvist, Calvin Tsay, and Ruth Misener.
\newblock Maximizing information from chemical engineering data sets: Applications to machine learning.
\newblock \emph{Chemical Engineering Science}, 252:\penalty0 117469, 2022.

\end{thebibliography}
